\documentclass{article}

\usepackage{amsmath,amsfonts,amssymb,times,graphicx,natbib,algorithm,algorithmic,hyperref}
\usepackage{slashbox,pict2e}
\newcommand{\keypoint}[1]{\vspace{0.1cm}\noindent\textbf{#1}\quad}

\usepackage[accepted]{whi2018}

\icmltitlerunning{Deep Neural Decision Trees}

\begin{document}

\twocolumn[
\icmltitle{Deep Neural Decision Trees}


\icmlsetsymbol{equal}{*}

\begin{icmlauthorlist}
\icmlauthor{Yongxin Yang}{uoe}
\icmlauthor{Irene Garcia Morillo}{uoe}
\icmlauthor{Timothy M. Hospedales}{uoe}
\end{icmlauthorlist}

\icmlaffiliation{uoe}{The University of Edinburgh}

\icmlcorrespondingauthor{Yongxin Yang}{yongxin.yang@ed.ac.uk}

\icmlkeywords{interpretability, transparency}

\vskip 0.3in
]


\printAffiliationsAndNotice{}

\begin{abstract}
Deep neural networks have been proven powerful at processing perceptual data, such as images and audio. However for tabular data, tree-based models are more popular. A nice property of tree-based models  is their natural interpretability. In this work, we present Deep Neural Decision Trees (DNDT) -- tree models realised by neural networks. A DNDT is intrinsically interpretable, as it is a tree. Yet as it is also a neural network (NN), it can be easily implemented in NN toolkits, and trained with gradient descent rather than greedy splitting. We evaluate DNDT on several tabular datasets, verify its efficacy, and investigate similarities and differences between DNDT and vanilla decision trees. Interestingly, DNDT self-prunes at both split and feature-level.


\end{abstract}

\section{Introduction}

The interpretability of predictive models is important, especially in cases where ethics are involved, such as law,
medicine, and finance; and mission critical applications where we wish to manually verify the correctness of a model's reasoning. Deep neural networks \cite{Yann2015,Schmidhuber2015} have achieved excellent performance in many areas, such as computer vision, speech processing, and language modelling. However  lack of interpretability prevents this family of black-box models from being used in applications for which we must know how the prediction is made in order to certify its decision process. Moreover, in some areas like Business Intelligence (BI), it is often more important to know how each factor contributes to the prediction rather than the conclusion itself. Decision tree (DT) based methods, such as C4.5 \cite{Quinlan1993} and CART \cite{Breimanbook}, have a clear advantage in this aspect, as one can easily follow the structure of the tree and check exactly how a prediction is made. 

In this work, we propose a new model at the intersection of these two approaches -- the Deep Neural Decision Tree (DNDT) -- and explore its connections to each. DNDTs are neural networks with a special architecture, where any setting of DNDT weights corresponds to a specific decision tree, and is therefore interpretable\footnote{The reverse is also true. For any DT, there is a corresponding DNDT that performs the same computation.}. 
However, as DNDT is realised by neural network (NN), it inherits several interesting properties different of conventional DTs: DNDT can be easily implemented in a few lines of code in any NN software framework; all parameters are simultaneously optimized with stochastic gradient descent rather than a more complex and potentially sub-optimal greedy splitting procedure; it is ready for large-scale processing with mini-batch-based learning and GPU acceleration out of the box, and it can be plugged into any larger NN model as a building block for end-to-end learning with back-propagation.

\section{Related Work}

\keypoint{Tree models}
Tree models are widely used in supervised learning, e.g., classification. They recursively partition the input space and assign a label/score to the final node. Well-known tree models include C4.5 \cite{Quinlan1993} and CART \cite{Breimanbook}. A key advantage of tree based models is that they are easy to interpret, since the predictions are given by a set of rules. It is also common to use an ensemble of multiple trees, such as Random Forest \cite{Breiman:2001:RF:570181.570182} and XGBoost \cite{Chen:2016:XST:2939672.2939785}, to boost performance at the expense of interpretability. Such tree-based models are often competitive or better than neural networks at predictive tasks using tabular data.

\keypoint{Model interpretability}
With machine learning based predictions becoming ubiquitous and affecting many aspects of our daily lives, the focus of research moves beyond model performance (e.g., efficiency and accuracy), to other factors such as interpretability \cite{Weller2017Challenges,DoshiKim2017Interpretability}. This is particularly so in applications where there are ethical \cite{bostrom_yudkowsky_2014} or safety concerns and models' predictions should be explainable in order to verify the correctness of their reasoning process or justify their decisions. 
There are now a number of attempts to make models explainable. Some are model-agnostic \cite{Ribeiro2016Lime}, while most are associated with a certain type of model, e.g., rule-based  classifiers \cite{7178589,Malioutov2017}, nearest neighbour models \cite{kim2016MMD}, and neural networks \cite{2017arXiv171111279K}.

\keypoint{Neural Networks and Decision Trees}
Some studies have proposed to unify neural network and decision tree models. \citet{6909412} proposed Neural Decision Forests (NDF) as an ensemble of neural decision trees, where the split functions are realised by randomized multi-layer perceptrons. Deep-NDF \cite{7410529} exploited a stochastic and differentiable decision tree model, which jointly learns the representations (via CNNs) and the classification (via decision trees). 
Our proposed DNDT differs from those methods in many ways. First, we do not have an alternative optimisation procedure for structure learning (splitting) and parameter learning (score matrix). Instead, we learn them all via back-propagation in a single pass. Second, we do not restrict that the splits to be binary (left or right), as we apply a differentiable binning function that can split nodes into multiple ($\ge 2$) leaves. Finally, and most importantly, we design our model specifically for interpretability, especially for application to tabular data, where we can interpret every input feature. In contrast, the models in \cite{6909412,7410529} are  designed for prediction performance and applied to raw image data. Some design decisions make them not appealing to tabular data. E.g., in \citet{7410529}, they use a less flexible tree where the structure is fixed while the node split is learned.

Despite the similar name, our work is fundamentally different to \citet{Balestriero2017Neural} which developed a kind of `oblique' decision tree realised by neural network. In contrast to conventional `univariate' decision trees, each node in their oblique decision tree involves \emph{all} features rather than a single feature, which renders the model uninterpretable. 

\keypoint{Alternative Decision Tree Inducers} Conventional DTs are learned by recursive greedy splitting of features \cite{Quinlan1993,Breimanbook}. This is efficient and has some benefits for feature selection, however such greedy search may be sub-optimal \cite{norouzi2015nonGreedyDT}. Some recent work explores alternative approaches to training decision trees which aim to achieve better performance with less myopic optimization, for example with latent variable structured prediction \cite{norouzi2015nonGreedyDT}, or training an RNN splitting controller using reinforcement learning \cite{Xiong2017Learning}.  In contrast, our DNDT is much simpler than these, but can still potentially find better solutions than conventional DT inducers by simultaneously searching the structure and parameters of the tree with SGD. Finally, we also note that while conventional DT inducers leverage only binary splits for simplicity, our DNDT model can equally easily work with splits of arbitrary cardinality, which can sometimes make for more interpretable trees.

\section{Methodology}

\subsection{Soft binning function}

The core module we implement here is a \emph{soft} binning function \cite{Dougherty95supervisedand} that we will use to make the split decisions in DNDT. Typically, a binning function takes as input a real scalar $x$ and produces an index of the bins to which $x$ belongs. Hard binning is non-differentiable, so we propose a differentiable approximation of this function.

Assuming we have a continuous variable $x$, that we want to bin into $n+1$ intervals. This leads to the need of $n$ cut points, which are trainable variables in this context. We denote the cut points as $[\beta_1, \beta_2, \dots, \beta_n]$ in a monotonically increasing manner\footnote{During training, the order of $\beta$'s may be shuffled up after updating, so we have to sort them first in every forward pass. However, this will not affect the differentiability because sort just swaps the positions of $\beta$'s.}, i.e., $\beta_1 < \beta_2 < \cdots < \beta_n$.

Now we construct a one-layer neural network with softmax as its activation function.
\begin{equation}
\label{eq1}
\pi = f_{w, b, \tau}(x) = \operatorname{softmax}((wx + b)/\tau)
\end{equation}
Here $w$ is a constant rather than a trainable variable, and its value is set as $w=[1,2,\dots,n+1]$. $b$ is constructed as,
\begin{equation}
\label{eq2}
b = [0, -\beta_1, -\beta_1 - \beta_2, \dots, -\beta_1 -\beta_2 - \cdots -\beta_n].
\end{equation}
and $\tau>0$ is a temperature factor. As $\tau \rightarrow 0$ the output tends to a one-hot vector.

We can verify it by checking three consecutive logits $o_{i-1}, o_{i}, o_{i+1}$. When we have both $o_{i} > o_{i-1}$ (so $x>\beta_i$) and $o_{i} > o_{i+1}$ (so $x<\beta_{i+1}$), $x$ must fall into the interval $(\beta_i, \beta_{i+1})$. Thus, the neural network in Eq.~\ref{eq1} will produce an \emph{almost} one-hot encoding of the binned $x$, especially with lower temperature. Optionally, we can apply the slope annealing trick \cite{Chung2016Hierarchical} that progressively reduces the temperature during training so that we can get a more deterministic model in the end.

If one prefers an \emph{actual} one-hot vector, Straight-Through (ST) Gumbel-Softmax \cite{Jang2016Categorical} can be applied: for the forward pass, we sample a one-hot vector using Gumbel-Max trick, while for the backward pass, we use Gumbel-Softmax to compute the gradient (see \citet{Bengio2013Estimating} for a more detailed analysis). 

\begin{figure}[t]
\centering
\includegraphics[width=0.23\textwidth]{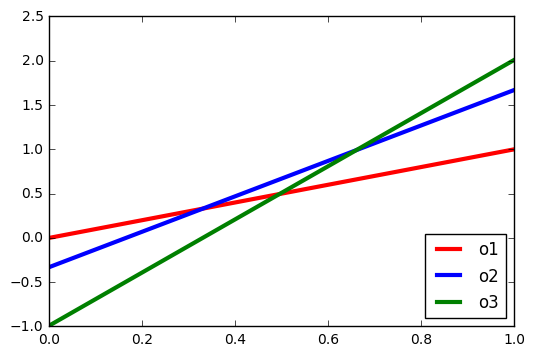}
\includegraphics[width=0.23\textwidth]{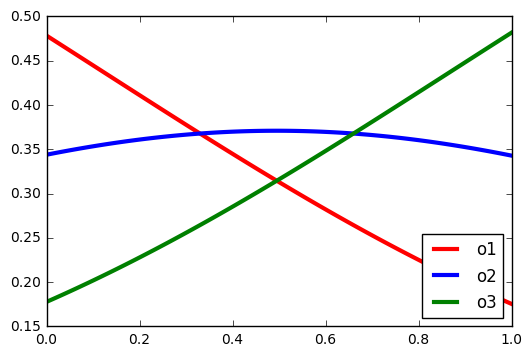}\\
\includegraphics[width=0.23\textwidth]{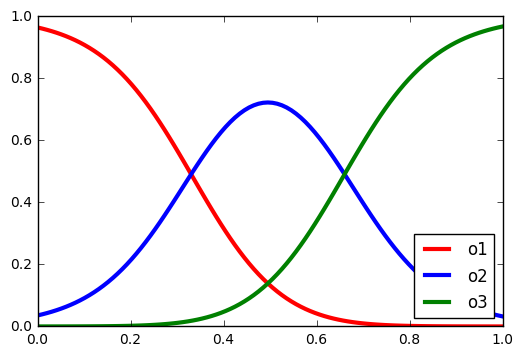}
\includegraphics[width=0.23\textwidth]{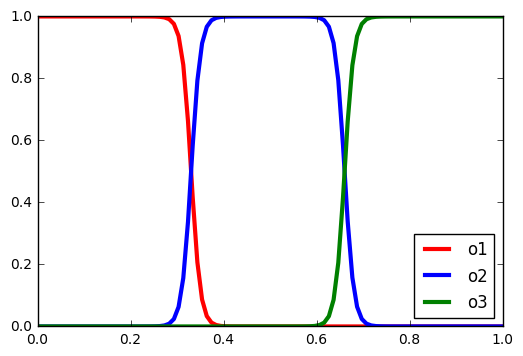}
\caption{A concrete example of our soft binning function using cut-points at $0.33$ and $0.66$. x-axis is the value of a continuous input variable $x\in [0,1]$. Top-left: the original values of logits; Top-right: values after applying softmax function with $\tau=1$; Bottom-left: $\tau=0.1$; Bottom-right: $\tau=0.01$.}
\label{fig1}
\end{figure}

Fig.~\ref{fig1} demonstrates a concrete example where we have a scalar $x$ in the range of $[0, 1]$ and two cut points at $0.33$ and $0.66$ respectively. Based on Eq.~\ref{eq1} and Based on Eq.~\ref{eq2}, we have the three logits $o_1 = x$, $o_2= 2x - 0.33$, $o_3 = 3x-0.99$.

\subsection{Making Predictions}
Given our binning function, the key idea is to construct the decision tree via Kronecker product $\otimes$. Assume we have an input instance $x\in\mathcal{R}^D$ with $D$ features. Binning each feature $x_d$ by its own neural network $f_{d}(x_d)$, we can exhaustively find all final nodes by,
\begin{equation}
z = f_1(x_1) \otimes f_2(x_2) \otimes \cdots \otimes f_D(x_{D}).
\end{equation}
Here $z$ is now also an almost one-hot vector that indicates the index of the leaf node where instance $x$ arrives. Finally, we assume a linear classifier at each leaf $z$ classifies instances arriving there.  DNDT is illustrated in Fig.~\ref{fig2}. 

\subsection{Learning the Tree} 
\begin{figure}[t]
\centering
\includegraphics[width=0.48\textwidth]{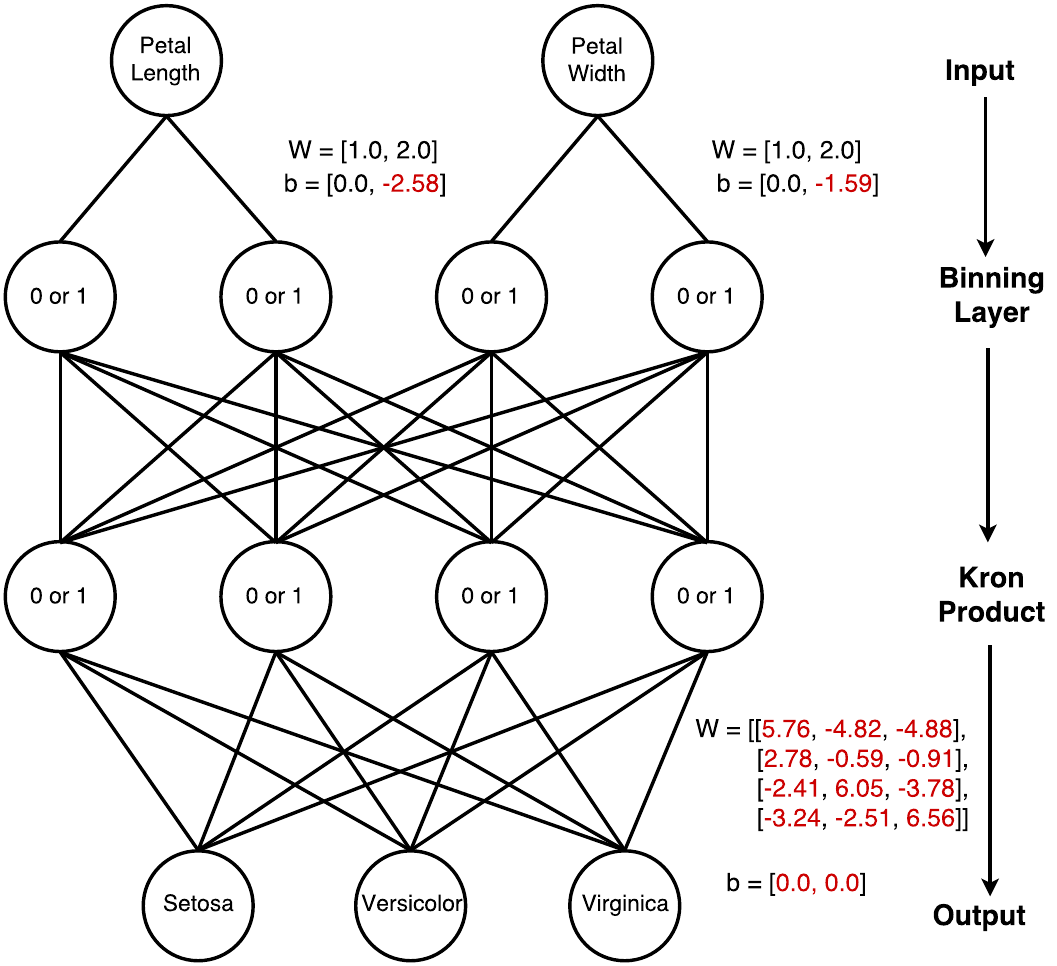}
\includegraphics[width=0.40\textwidth]{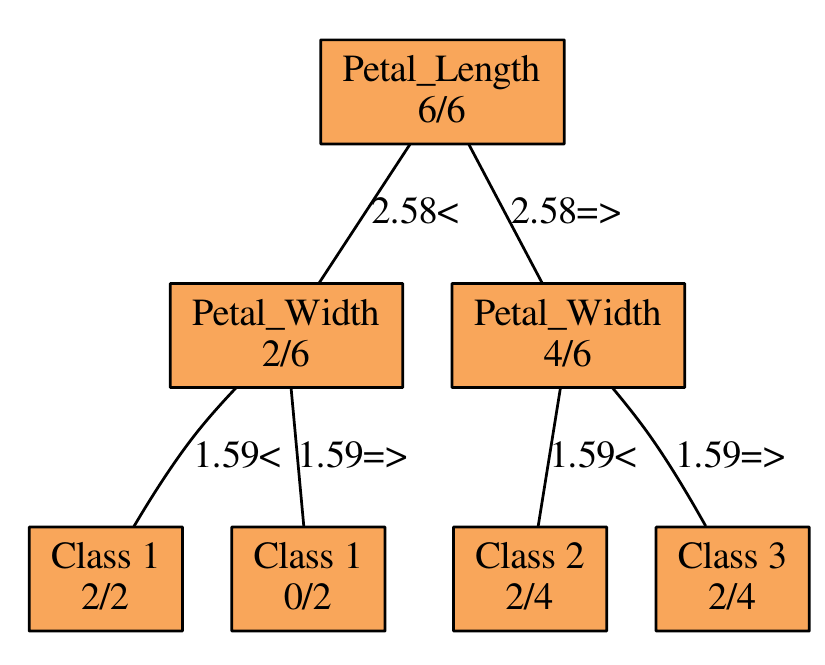}
\caption{A learned DNDT for the \emph{Iris} dataset (reduced two-feature version). Top: DNDT-view where {red fonts} indicate trainable variables, and black indicates constants. Below: DT-view. The same network rendered as a conventional decision tree. The fractions indicate the route of a randomly chosen $6$ instances being classified.}
\label{fig2}
\end{figure}

With the method described so far we can route input instances to leaf nodes and classify them. Thus training a decision tree now becomes a matter of training the bin cut points and leaf classifiers. Since all steps of our forward pass are differentiable, all parameters (Fig.~\ref{fig2}, red) can now be straightforwardly and simultaneously trained with SGD.

\paragraph{Discussion}
DNDT scales well with number of instances due to neural network style mini-batch training. However a key drawback of the design so far is that, due to the use of Kronecker product, it is not scalable with respect to the number of features. In our current implementation, we avoid this issue with `wide' datasets by training a forest with random subspace \cite{709601} -- at the expense of our interpretibility. That is, introducing multiple trees, each trained on a random subset of features. 
A better solution that does not require  an uninterpretable forest is to exploit the sparsity of the final binning during learning: the number of non-empty leaves grows much slower than the total number of leaves. But this somewhat complicates the otherwise simple implementation of DNDT.

\section{Experiments}

\subsection{Implementation}
DNDT is conceptually simple and easy to implement with $\approx 20$ lines code in TensorFlow \cite{tensorflow} or PyTorch \cite{paszke2017automatic}\footnote{\url{https://github.com/wOOL/DNDT}}. Because it is implemented as a neural network, DNDT supports `out of the box' GPU acceleration and mini-batch based learning of datasets that do not fit in memory, thanks to modern deep learning frameworks. 

\subsection{Datasets and Competitors}
We compare DNDT against neural networks (implemented by TensorFlow \cite{tensorflow}) and decision tree (from Scikit-learn \cite{scikit-learn}) on $14$ datasets collected from Kaggle and UCI (see Tab.~\ref{tab1} for dataset details).

\begin{table}[t]
\centering
\begin{tabular}{l l l l}
\hline 
Dataset & $\#$inst. & $\#$feat. & $\#$cl.\\
\hline
Iris & 150 & 4 & 3\\
Haberman's Survival & 306 & 3 & 2\\
Car Evaluation & 1728 & 6 & 4\\
Titanic (K) & 714 & 10 & 2\\
Breast Cancer Wisconsin & 683 & 9 & 2\\
Pima Indian Diabetes (K) & 768 & 8 & 2\\
Gime-Me-Some-Credit (K) & 201669 & 10 & 2\\
Poker Hand & 1025010 & 11 & 9\\
Flight Delay & 1100000 & 9 & 2\\
HR Evaluation (K) & 14999 & 9 & 2\\
German Credit Data & 1000 & 20 & 2\\
Connect-4 & 67557 & 42 & 2\\
Image Segmentation & 2310 & 19 & 7\\
Covertype & 581012 & 54 & 7\\
\hline 
\end{tabular}
\caption{Collection of 14 datasets from Kaggle (indicated with (K)) and UCI: number of instances ($\#$inst.), number of features ($\#$feat.), and number of classes ($\#$cl.)}
\label{tab1}
\end{table}

For decision tree (DT) baseline we set two of the key hyper-parameters \emph{criterion} as `gini' and \emph{splitter} as `best'. For neural network (NN), we use an architecture of two hidden layers with 50 neurons each for all datasets. DNDT also has a hyper-parameter, the number of cut points for each feature (branching factor),  which we set to 1 for all features and datasets. A detailed analysis of the effect of this hyper-parameter can be found in Sec.~\ref{sec1}. For datasets with more than $12$ features, we use an ensemble of DNDT, where each tree picks $10$ features randomly, and we have $10$ tress in total. The final prediction is given by majority voting.

\subsection{Accuracy}

We evaluate the performance of DNDT, decision tree, and neural network models on each of the datasets in Tab.~\ref{tab1}. The test set accuracies are presented in Tab.~\ref{tab2}. 

\begin{table}[t]
\centering
\begin{tabular}{l l l l}
\hline 
Dataset & DNDT & DT & NN\\
\hline
Iris & \textbf{100.0} & \textbf{100.0} & \textbf{100.0}\\
Haberman's Survival & \textbf{70.9} & 66.1 & \textbf{70.9}\\
Car Evaluation & 95.1 & \textbf{96.5} & 91.6\\
Titanic & \textbf{80.4} & 79.0 & 76.9\\
Breast Cancer Wisconsin & 94.9 & 91.9 & \textbf{95.6}\\
Pima Indian Diabetes & 66.9 & \textbf{74.7} & 64.9\\
Gime-Me-Some-Credit & 98.6 & 92.2 & \textbf{100.0}\\
Poker Hand & 50.0 & \textbf{65.1} & 50.0\\
Flight Delay & \textbf{78.4} & 67.1 & 78.3\\
HR Evaluation & 92.1 & \textbf{97.9} & 76.1\\
German Credit Data (*) & \textbf{70.5} & 66.5 & \textbf{70.5}\\
Connect-4 (*) & 66.9 & \textbf{77.7} & 75.7\\
Image Segmentation (*) & 70.6 & \textbf{96.1} & 48.05\\
Covertype (*) & 49.0 & \textbf{93.9} & 49.0\\
\hline
$\#$ of wins & 5 & \textbf{7} & 5\\
Mean Reciprocal Rank & 0.65 & \textbf{0.73} & 0.61\\
\hline 
\end{tabular}
\caption{Test set accuracy of each model: DT: Decision tree.  NN:  neural network. DNDT: Our deep neural decision tree, where (*) indicates that the ensemble version is used.}
\label{tab2}
\end{table}

Overall the best performing model is the DT. DT's good performance is not surprising because these datasets are mainly tabular and the feature dimension is relatively low. Conventionally, neural networks do not have a clear advantage on this kind of data. However, DNDT is slightly better than the vanilla neural network, as it is closer to decision tree by design. Of course this is only an indicative result, as all of these models have tuneable hyperparameters.
Nevertheless, it's interesting that no model has a dominant advantage. This is reminiscent of \emph{no free lunch theorems} \cite{Wolpert1966Lack}.

\subsection{Analysis of active cut-points}
\label{sec1}
In DNDT the number of cut points per feature is the model complexity parameter. We do not bound the cut points' values, which means it is possible that some of them are inactive. E.g., they are either smaller than the minimal $x_d$ or greater than the maximal $x_d$.

\begin{figure}[t]
\centering
\includegraphics[width=0.48\textwidth]{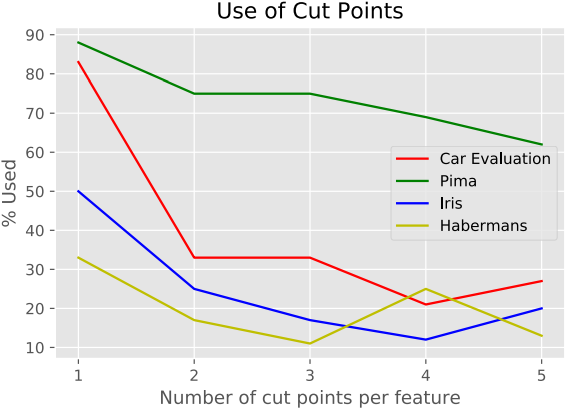}
\caption{Percentage (\%) of active cut points used by DNDT.}
\label{fig3}
\end{figure}

In this section, we investigate how many of cut points are \emph{actually} used after DNDT learning. A cut point is active when at least one instance from the dataset falls on each side of it. 
For four datasets, Car Evaluation, Pima, Iris, and Haberman's, we set the number of cut points per feature from 1 to 5, and calculate the percentage of active cut points, as shown in Fig.~\ref{fig3}. We can see that as the number of cut points increases, their utilisation generally decreases. This implies that DNDT is somewhat self-regularising: it does not make use of all the parameters available to it.

\begin{figure}[t]
\centering
\includegraphics[width=0.23\textwidth]{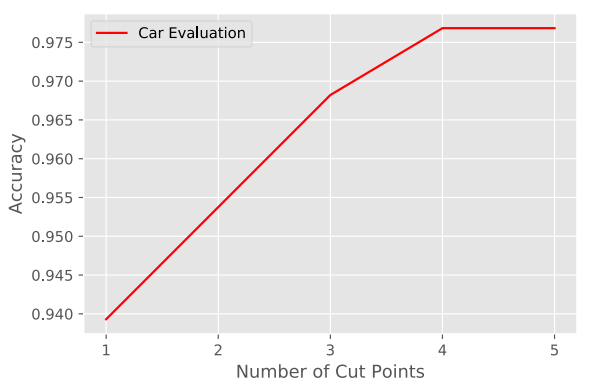}
\includegraphics[width=0.23\textwidth]{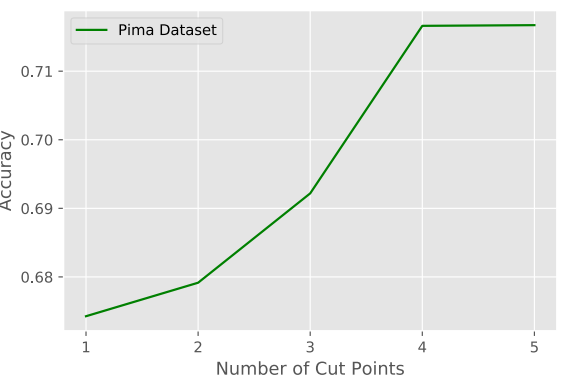}\\
\includegraphics[width=0.23\textwidth]{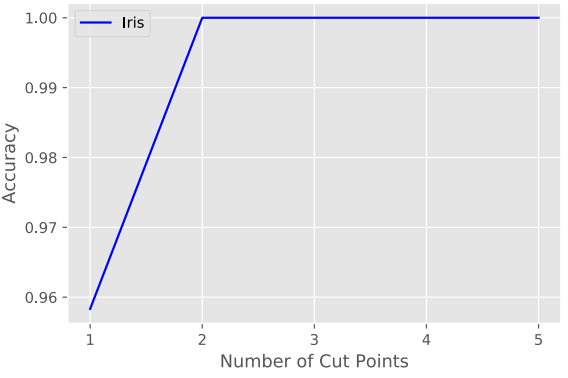}
\includegraphics[width=0.23\textwidth]{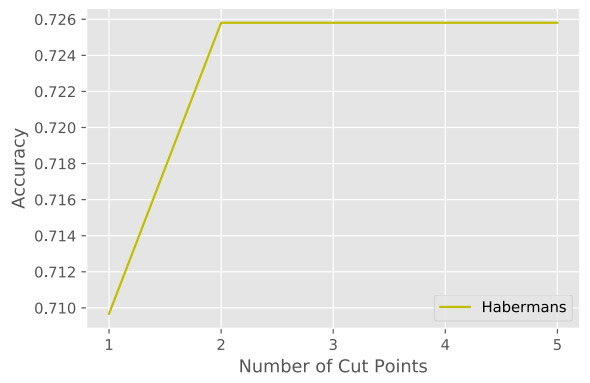}
\caption{Test accuracy of DNDT for increasing number of cut points (model complexity).}
\label{fig4}
\end{figure}

We can further investigate how the number of available cut points affects performance on these datasets. As we can see in Fig.~\ref{fig4},  performance initially increases with more cut points, before stabilising after a certain value. This is reassuring because it means that large DNDTs do not over-fit the training data, even without explicit regularisation.

\subsection{Analysis of active features}
\label{sec2}

In DNDT learning, it is also possible that for a certain feature all cut points are inactive.  This corresponds to disabling the feature, so that it does not impact prediction. It is analogous to a conventional DT learner never selecting a given feature to make a split anywhere in the tree. In this section we analyse how DNDT rules out features in this way.  We run DNDT 10 times, and record the number of times a given feature is excluded because all its cut points are inactive.

Given randomness  from both weight initialisation and mini-batch sampling, we  observe that some features (e.g., index 0 feature in iris) are consistently ignored by DNDT (See Tab.~\ref{tab3} for all results). This suggests that DNDT does some implicit feature selection by pushing cut points out of the data boundary for unimportant features. As a side product, we can obtain a measure of feature importance from  feature selection over  multiple runs: The more times a feature is ignored, the less important it is likely to be.

\begin{table}[t]
\centering
\resizebox{0.48\textwidth}{!}{
\begin{tabular}{l | l l l l l l l l l l}
\hline 
\backslashbox{Dataset}{Feat. Idx} & 0 & 1 & 2 & 3 & 4 & 5 & 6 & 7 & 8 & 9 \\
\hline
Haberman's & \textbf{100} & \textbf{100} & 0 &  - & - & - & - & - & - & - \\
Iris & \textbf{100} & 90 & 50 & 10 &  - & - & - & - & - & - \\
Pima & 10 & 0 & 0 & 0 & 20 & 0 & 0 & \textbf{100} & - & - \\
Titanic & 0 & 0 & 0 & 0 & 0 & 10 & 20 & 10 & 20 & 40\\
\hline 
\end{tabular}
}%
\caption{Percentage (\%) of times that DNDT ignores each feature.}
\label{tab3}
\end{table}

\subsection{Comparison to decision tree}

\begin{figure}[h]
\centering
\includegraphics[width=0.23\textwidth]{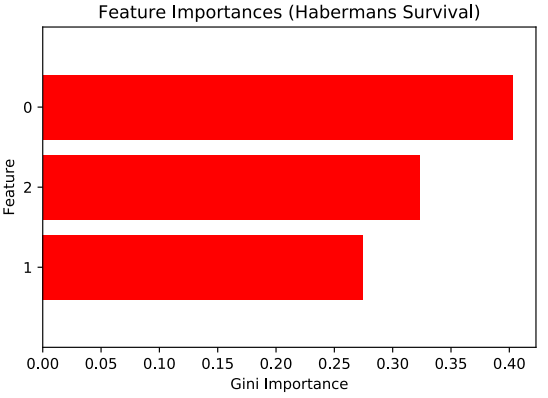}
\includegraphics[width=0.23\textwidth]{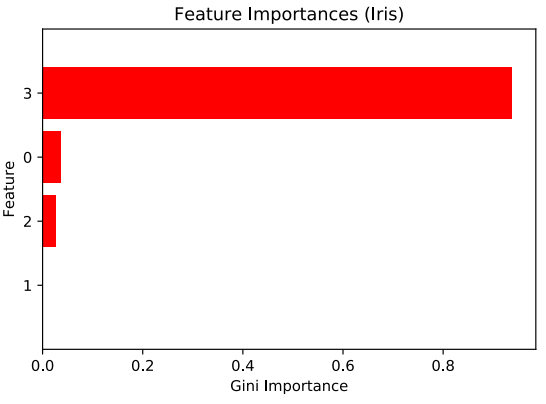}\\
\includegraphics[width=0.23\textwidth]{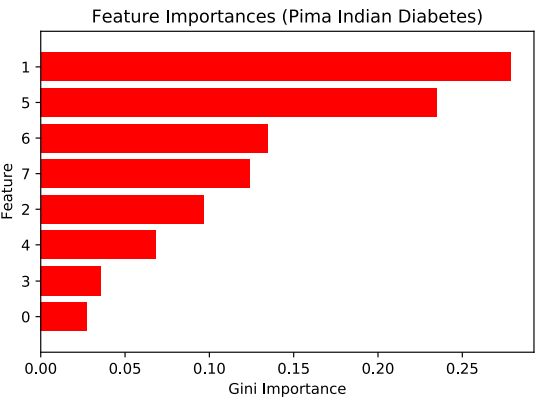}
\includegraphics[width=0.23\textwidth]{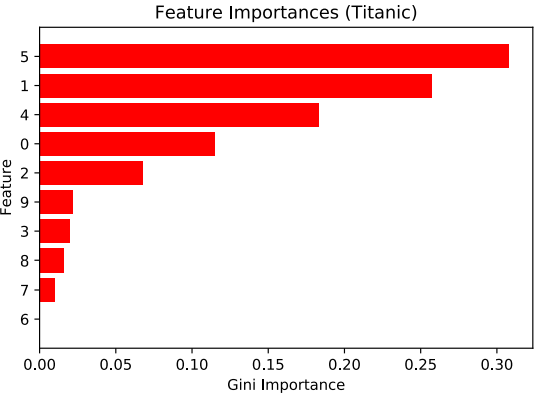}
\caption{Feature importance ranking produced by DT (Gini).}
\label{fig5}
\end{figure}

Using the techniques developed in Sec.~\ref{sec2}, we investigate whether DNDT and DT favour similar features. We compare the the feature importance through Gini used in decision tree (Fig.~\ref{fig5}) with our  selection rate metric (Table~\ref{tab3}).

Comparing these results we see that sometimes DNDT and DT share a feature selection  preference. E.g., for Iris, they both rank feature 3 as the most important. But it happens that they can also have different views, e.g., for Haberman's, DT picked feature 0 as the most important, whereas DNDT completely ignored it. In fact, DNDT only makes use of feature 2  for prediction, which is ranked second by DT. However, this kind of disagreement may not necessarily lead to significantly different performance. As we can see in Tab.~\ref{tab2}, for Haberman's, the test accuracies of DNDT and DT are 70.9\% and 66.1\% respectively.

Finally, we quantify the similarity between DNDT feature ranking and DT feature ranking by calculating Kendall's Tau of two ranking lists. The results  in Tab.~\ref{tab4}  suggest a moderate correlation overall. 
\begin{table}[t]
\centering

\begin{tabular}{l | l l l l}
\hline 
Dataset & Titanic & Iris & Pima & Habermans\\
Kendall's Tau & 0.4 & 0.33 & 0.32 & 0.0\\
\hline 
\end{tabular}
\caption{Kendall's Tau of DNDT's and DT's feature ranking: larger values mean `more similar'}
\label{tab4}
\end{table}


\subsection{GPU Acceleration}

\begin{figure}[h]
\centering
\includegraphics[width=0.48\textwidth]{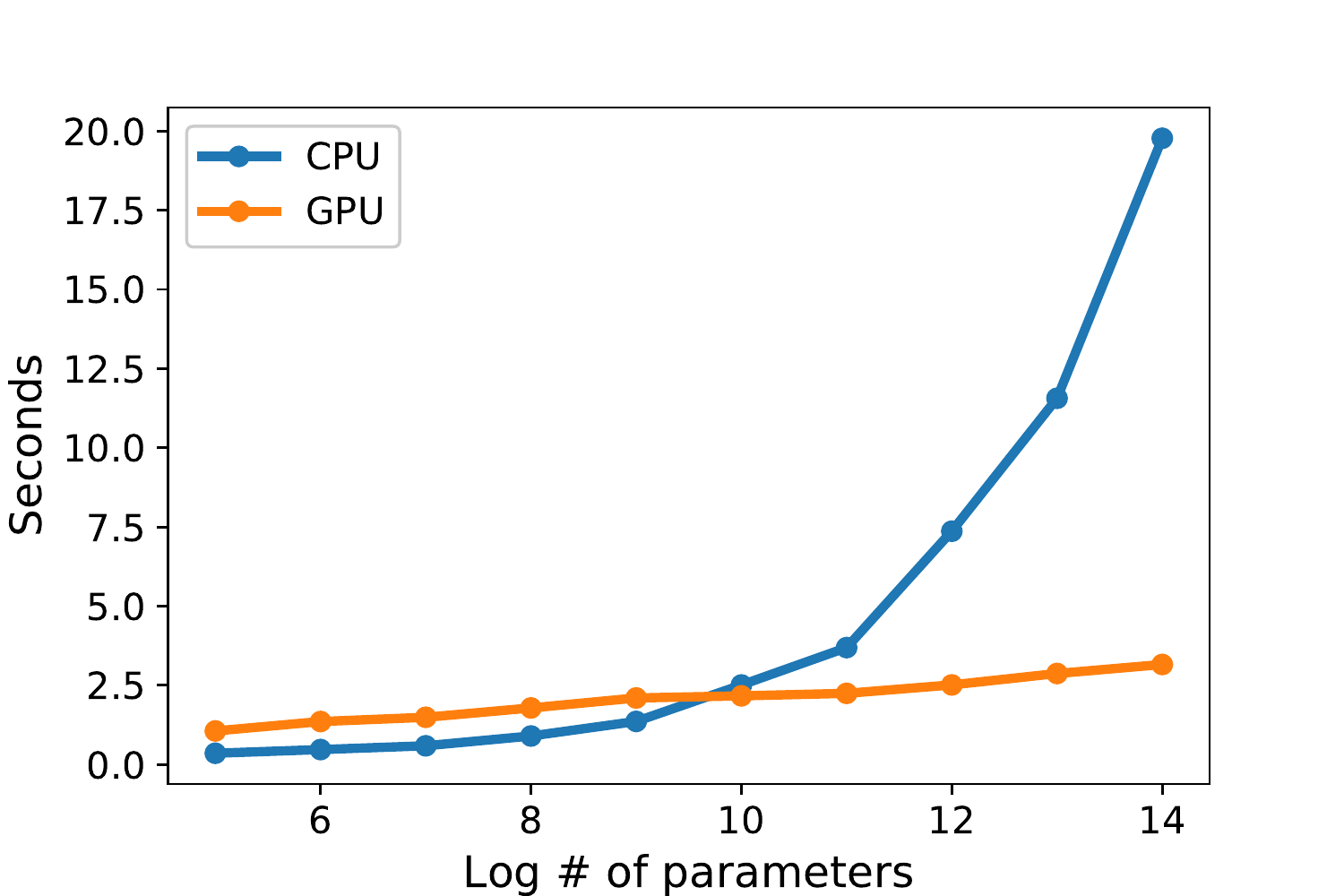}
\caption{GPU Acceleration illustration: DNDT training time on 3.6GHz CPU vs GTX Titian GPU. Average over 5 runs.}
\label{fig6}
\end{figure}

Finally we verify the ease of accelerating DNDT learning of DTs by GPU processing -- a capability not common or straightforward for conventional DT learners. By increasing the number of cut points for each feature, we can get larger models, for which GPU mode has significantly shorter running time (see Fig.~\ref{fig6}). 

\section{Conclusion}
We introduced a neural network based tree model DNDT. It has better performance than NNs for certain tabular datasets, while providing an interpretable decision tree. Meanwhile compared to conventional DTs, DNDT is simpler to implement, simultaneously searches tree structure and parameters with SGD, and is easily GPU accelerated.

There are many avenues for future work. We want to investigate the source of self-regularisation that we observed; explore plugging in DNDT as a module connected to a conventional CNN feature learner for end-to-end learning; find out whether DNDT's whole-tree SGD-based learning can be used as postprocessing to fine-tune conventional greedily trained DTs and improve their performance; and  find out whether the many NN-based approaches to transfer learning can be leveraged to enable transfer learning for DTs.

\textbf{Acknowledgements} This work was supported by the EPSRC grant EP/R026173/1.


\end{document}